\documentclass[11pt,table,a4paper,nohyperref]{article}
\usepackage{style/acl2020}

\usepackage{xcolor}

\newcommand{\figfile}[1]{2020_lrec_sense/figures/#1}

\newcommand{\appendixmention}[1]{#1}
\newcommand{\gasi}{\abr{gasi}}
\newcommand{\wic}{\abr{w}{\small i}\abr{c}}
\newcommand{\sasi}{\abr{sasi}}

\newif\ifcomment\commentfalse

\usepackage[a-1b]{pdfx}

\usepackage{framed}
\usepackage{mdwlist}
\usepackage{latexsym}
\usepackage{colortbl}
\usepackage{xcolor}
\usepackage{nicefrac}
\usepackage{booktabs}
\usepackage{amsfonts}
\usepackage[T1]{fontenc}
\usepackage{bold-extra}
\usepackage{amsmath}
\usepackage{amssymb}
\usepackage{bm}
\usepackage{graphicx}
\usepackage{mathtools}
\usepackage{microtype}
\usepackage{multirow}
\usepackage{multicol}
\usepackage{latexsym,comment}
\usepackage[normalem]{ulem}

\newcommand*{\missingreference}{{\Huge \colorbox{red}{?reference?}}}
\newcommand*{\missingcitation}{{\Huge \colorbox{red}{?citation?}}}
\makeatletter
\def\@setref#1#2#3{%
   \ifx#1\relax
    \protect\G@refundefinedtrue
    \nfss@text{\reset@font\missingreference}%
    \@latex@warning{Reference `#3' on page \thepage \space
              undefined}%
   \else
    \expandafter#2#1\null
   \fi}
\def\@citex[#1]#2{\leavevmode
   \let\@citea\@empty
   \@cite{\@for\@citeb:=#2\do
     {\@citea\def\@citea{,\penalty\@m\ }%
      \edef\@citeb{\expandafter\@firstofone\@citeb\@empty}%
      \if@filesw\immediate\write\@auxout{\string\citation{\@citeb}}\fi
      \@ifundefined{b@\@citeb}{\hbox{\reset@font\missingcitation}%
        \G@refundefinedtrue
        \@latex@warning
          {Citation `\@citeb' on page \thepage \space undefined}}%
        {\@cite@ofmt{\csname b@\@citeb\endcsname}}}}{#1}}
\makeatother

\newcommand{\gem}[1]{\mbox{\textsc{gem}}}
\newcommand{\abr}[1]{\textsc{#1}}

\newcommand{\explain}[2]{\underbrace{#2}_{\mbox{\footnotesize{#1}}}}

\newcommand{\g}{\, | \,}

\DeclareMathOperator*{\argmax}{arg\,max}

\newcommand{\bmat}[1]{\text{\textbf{#1}}}
\newcommand{\bvec}[1]{\boldsymbol{#1}}

\newcommand{\hidetext}[1]{}
\newcommand{\ignore}[1]{}

\ifcomment
\newcommand{\pinaforecomment}[3]{\colorbox{#1}{\parbox{.8\linewidth}{#2: #3}}}
\else
\newcommand{\pinaforecomment}[3]{}
\fi

\newcommand{\jbgcomment}[1]{\pinaforecomment{red}{JBG}{#1}}

\newcommand{\smallurl}[1]{ \begin{tiny}\url{#1}\end{tiny}}

\definecolor{lightblue}{HTML}{3cc7ea}
\definecolor{CUgold}{HTML}{CFB87C}
\definecolor{grey}{rgb}{0.95,0.95,0.95}
\definecolor{ceil}{rgb}{0.57, 0.63, 0.81}
\definecolor{UMDred}{HTML}{ed1c24}
\definecolor{UMDyellow}{HTML}{ffc20e}

\newcommand{\muse}{\textsc{muse}}

\newcommand{\elmo}{\textsc{elm}{\small o}}

\newcommand{\glove}{\abr{gl}{\small o}\abr{ve}}

\usepackage{cleveref}

\DeclareMathOperator{\softmax}{softmax}
\DeclareMathOperator{\onehot}{one\_hot}
\DeclareMathOperator{\mean}{mean}

\newcommand{\secref}[1]{Sec.~\ref{#1}}

\aclfinalcopy

	\title{Which Evaluations Uncover Sense Representations that Actually Make Sense?}

                \author{Fenfei Guo, Jordan Boyd-Graber, Mohit Iyyer, Leah Findlater}

\begin{document}
\maketitle


\begin{abstract}
  Text representations are critical for modern natural language processing. One form of text representation, sense-specific embeddings, reflect a word's sense in a sentence better than single-prototype word embeddings tied to each type. However, existing sense representations are not uniformly better: although they work well for computer-centric evaluations, they fail for human-centric tasks like inspecting a language's sense inventory. To expose this discrepancy, we propose a new coherence evaluation for sense embeddings. We also describe a minimal model (Gumbel Attention for Sense Induction) optimized for discovering interpretable sense representations that are more coherent than existing sense embeddings. \\
\end{abstract}

        \jbgcomment{Cite:
https://www.aclweb.org/anthology/D19-1007.pdf

https://arxiv.org/abs/1908.05646

https://www.aclweb.org/anthology/D19-1005/

Don't cite this paper, but cite the {\bf supervised} embeddings in here:
https://www.aclweb.org/anthology/D19-1009.pdf
}

\section{Context, Sense, and Representation}

Computers need to represent the meaning of words in context.
\abr{bert}~\cite{devlin-18} and \elmo{}~\cite{peters-18} have
dramatically changed how natural language processing represents text.
Rather than one-size-fits-all word vectors that ignore the nuance of
how words are used in context, these new representations have 
topped the leaderboards for question answering, inference, and
classification.

Contextual representations have supplanted 
multisense embeddings~\cite{CamachoCollados-18}.
While these methods learn a vector for \emph{each sense}, they do not work
encode meanings in downstream tasks as well as contextual representations~\cite{peters-18}.

However, computers are not the only consumer of text representations.
Humans also use word representations to understand diachronic drift,
investigate a language's sense inventory, or to cluster and explore
documents.  Thus, a primary role for multisense word
embeddings is \emph{human} understanding of word meanings.  Unfortunately, multisense models have only been evaluated on
\emph{computer}-centric dimensions and have ignored the question of
\emph{sense interpretability}.  

We first develop measures for how well models encode and explain a
word's meaning to a human (\secref{sec:intp}).
Existing multisense models do not necessarily fare best on this
evaluation; our simpler model (Gumbel Attention for Sense Induction:
\gasi{}, \secref{sec:model}) that focuses on
discrete sense selection can better capture human-interpretable
representations of senses; comparing against
traditional evaluations (\secref{sec:wordsim}), \gasi{}
has better contextual word similarity and 
competitive non-contextual word similarity.
Finally, we discuss the connections between representation
learning and how modern contextual representations could 
better capture interpretable senses (\secref{sec:related}).

\begin{figure}[t]
  \includegraphics[width=0.95\linewidth]{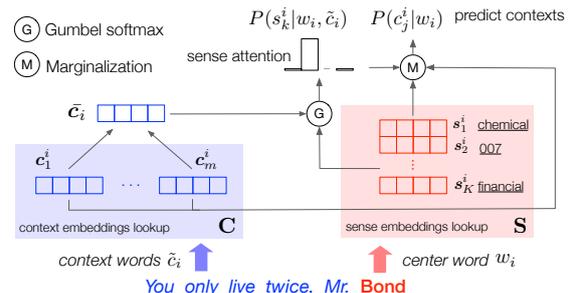}
  \vspace{-0.5cm}
  \caption{Network structure with an example of our \gasi{} model which
    learns a set of global context embeddings $\bmat{C}$ and a set of
    sense embeddings $\bmat{S}$.}\label{fig:gasi_net}
\vspace{-0.5cm}
\end{figure}

\ignore{-0.3cm}
\section{Attentional Sense Induction}
\label{sec:model}

Before we explore human interpretability of sense induction, we first
describe our simple models to disentangle word senses.  Our two models
are built on Word2Vec~\citep{mikolov2013a,mikolov2013b}, which we review in \secref{sec:skip-gram}.  Both
models use a straightforward attention mechanism to select which sense
is used in a token's context, which we contrast to alternatives for
sense selection (\secref{sec:attention-muse}).
Building on these foundations, we introduce our model, \gasi{},
and along the way introduce a soft-attention stepping-stone (\sasi{}).

\ignore{-0.3cm}
\subsection{Foundations: Skip-Gram and Gumbel}
\label{sec:skip-gram}

Word2Vec
jointly learns word embeddings~$\bmat{W} \in \mathbb{R}^{|V|\times d}$
and context embeddings~$\bmat{C} \in \mathbb{R}^{|V|\times d}$. More specifically, given a
 vocabulary~$V$ and  embedding dimension~$d$, it
maximizes the likelihood of the context words~$c_{j}^i$ that
surround a given center word~$w_i$ in a context window~$\tilde{c}_i$,
\begin{equation}
 J(\bmat{W}, \bmat{C}) \propto \sum_{w_i \in V}\sum_{c_{j} ^i\in \tilde{c}_i} \log P(c_{j}^i \g w_i; \bmat{W}, \bmat{C}),
\label{eq:obj_sg}
\end{equation}
where $P(c_{j}^i\g w_i)$ is over the vocabulary,
 \begin{equation}
 	P(c_{j}^i \g w_i; \bmat{W}, \bmat{C}) = \frac{ \exp \left({\bvec{c}_j^i} ^{\top} \bvec{w}_i\right)}{\sum_{c \in V}  \exp \left(\bvec{c}^{\top}\bvec{w}_i\right)}.
 \label{eq:sg_softmax}
 \end{equation}
In practice, $\log P(c_{j}^i \g w_i)$ is approximated by
negative sampling.  We extend it to learn
representations for individual word senses.

\subsection{Gumbel Softmax}
\label{sec:gs}

As we introduce word senses, our model will need to select
\emph{which} sense is relevant for a context.  The Gumbel
softmax~\citep{jang2016categorical,maddison2016concrete} approximates
the sampling of discrete random variables; we use it to select the
sense.  Given a discrete random variable $X$ with
$P(X=k) \propto \alpha_k$, $\alpha_k \in (0, \infty)$, the
Gumbel-max~\citep{gumbel1954statistical}
refactors the sampling of $X$ into
\begin{eqnarray}
X = \argmax_k (\log \alpha_k + g_k),
\end{eqnarray}
where the Gumbel noise $g_k = -\log(-\log(u_k))$ and $u_k$ are i.i.d.
 from Uniform(0, 1).
The Gumbel softmax approximates sampling $
\onehot(\argmax_k (\log \alpha_k + g_k))$ by
 \begin{equation}
y_k = \softmax((\log \alpha_k  + g_k)/\tau).
 \end{equation}
Unlike soft selection of senses, the Gumbel softmax can make harder
selections, which will be more interpretable to humans.





\jbgcomment{  Reframe this section as a review of existing models
  (esp. MUSE) and introducing your variant.  }



\subsection{Why Attention?  Musing on Alternatives}
\label{sec:attention-muse}

For fine-grained sense inventories, it makes sense to have graded
assignment of tokens to senses~\cite{erk-09,jurgens-15}.  However, for
coarse senses---except for humor~\cite{miller-17}---words
typically are associated with a \emph{single sense}, often a single
sense per discourse~\cite{gale-92}.  A good model should respect this.
Previous models either use non-differentiable objectives or---in the
case of the current state of the art,
\muse{}~\cite{Muse}---reinforcement learning to select word senses.
By using Gumbel softmax, our model both approximates discrete sense
selection and is differentiable.

As we argue in the next section, applications with a human in the loop
are best assisted by discrete senses; the Gumbel softmax, which we
develop for our task here, helps us discover these discrete senses.

\subsection{Attentional Sense Induction}
\label{sec:sasi}


\paragraph{Embeddings} 

We learn a context embedding matrix~$\bmat{C} \in
\mathbb{R}^{|V|\times d}$ and a sense embedding tensor~$\bmat{S} \in
\mathbb{R}^{|V|\times K \times d }$. Unlike previous
work~\citep{neelakantan2015efficient,Muse}, no extra embeddings are
kept for sense induction.

\ignore{-0.2cm}
\paragraph{Number of Senses}

For simplicity and consistency with previous work, our model has
$K$ fixed senses. Ideally, if we set a large number of $K$, with a perfect
pruning strategy, we can estimate the number of senses per type by removing
duplicated senses.

However, this is challenging~\cite{mccarthy-16}; 
instead we use a simple pruning strategy.
We estimate a pruning threshold $\lambda$ by averaging 
the estimated duplicate sense and true neighbor distances,
\begin{equation}
\lambda = \frac{1}{2}(\mean(D_{dup}) + \mean(D_{nn})),
\label{eq:dup}
\end{equation}
where $D_{dup}$ are the cosine distances for duplicated sense pairs and $D_{nn}$ is
that of true neighbors (different types). We sample 100 words and 
if two senses are top-5 nearest neighbors of each other, we consider them duplicates.

After pruning duplicated senses with $\lambda$, we can retrain a new model with
estimated number of senses for each type by masking the sense attentions.\appendixmention{\footnote{More details and analysis about pruning are in Appendix C}} Results
in Table~\ref{tab:human} and~\ref{tab:wsim} validate our pruning strategy.

\ignore{-0.2cm}
\paragraph{Sense Attention in Objective Function}

Assuming a center word~$w_i$ has senses $\{s_1^i, s_2^i, \dots,s_K^i
\}$, the original Skip-Gram likelihood becomes a marginal
distribution over all senses of $w_i$ with sense induction
probability $P(s_k^i \g w_{i})$; we focus on the disambiguation
given local context $\tilde{c}_i$ and estimate $P(s_k^i \g w_{i})  \approx P(s_k^i \g w_{i}, \tilde{c}_i)$;
and thus,
\begin{equation}
  P(c_j^i  \g w_i) \approx \sum_{k=1}^{K} P(c_j^i  \g s_k^i) \explain{attention}{P(s_k^i \g w_{i}, \tilde{c}_i)},
\label{eq:dis}
\end{equation}

Replacing $P(c_j^i \g w_i)$ in Equation~\ref{eq:obj_sg} with
Equation~\ref{eq:dis} gives our objective function $J(\bmat{S},
\bmat{C}) \propto$
\begin{equation}
 \sum_{w_i \in V}\sum_{c_j^i \in \tilde{c}_i} \log
\sum_{k=1}^{K} P(c_j^i  \g s_k^i)P(s_k^i \g w_{i}, \tilde{c}_i).
\label{eq:jatt}
\end{equation}

\ignore{-0.2cm}
\paragraph{Modeling Sense Attention}

We can model the \emph{contextual sense induction
  distribution} with soft attention; we call the resulting model
soft-attention sense induction (\sasi{}); although it is a stepping
stone to our final model, we compare against it in our experiments as
it isolates the contributions of hard attention.  In \sasi{}, the
sense attention is conditioned on the entire local context
$\tilde{c}_i$ with softmax:
\begin{equation}
P(s_k^i \g w_{i}, \tilde{c}_i) = \frac{\exp \left(\bar{\boldsymbol{c}}_i^{\top}\boldsymbol{s}_k^i\right)}{\sum_{k=1}^{K}\exp \left(\bar{\boldsymbol{c}}_i^{\top}\boldsymbol{s}_k^i \right)},
\label{eq:sasi}
\end{equation}
where $\bar{\bvec{c}}_i$ is the mean of the context vectors in
$\tilde{c}_i$.  \appendixmention{A derivation of how this affects negative sampling is
in Appendix~\ref{sec:deriv-dess}.}

\ignore{-0.2cm}
\subsection{Scaled Gumbel Softmax for Sense Disambiguation}
\label{sec:sgs}



To learn \emph{distinguishable sense
  representations}, we implement \emph{hard} attention in our full
model, Gumbel Attention for Sense Induction (\gasi{}).
While hard attention is conceptually attractive, it can increase
computational difficulty: discrete choices are not differentiable and
thus incompatible with modern deep learning frameworks.
To preserve differentiability (and resorting to equally complex reinforcement
learning), we apply the
 Gumbel softmax reparameterization trick to
our sense attention function (Equation~\ref{eq:sasi}).
 
 
\paragraph{Vanilla Gumbel}

The discrete sense sampling from Equation~\ref{eq:sasi} can be
refactored 
\begin{equation}
\bvec{z^i} = \onehot(\argmax_k (\bar{\boldsymbol{c}_i}^{\top}\boldsymbol{s}_k^i + g_k)),
\label{eq:oh}
\end{equation}
and the hard attention approximated 
\begin{equation}
y_k^i = \softmax ((\bar{\boldsymbol{c}_i}^{\top}\boldsymbol{s}_k^i  + g_k)/\tau).
\label{eq:gs}
\end{equation}

\paragraph{Scaled Gumbel}

\begin{figure}[t]
	\centering
	\includegraphics[width=1.0\linewidth]{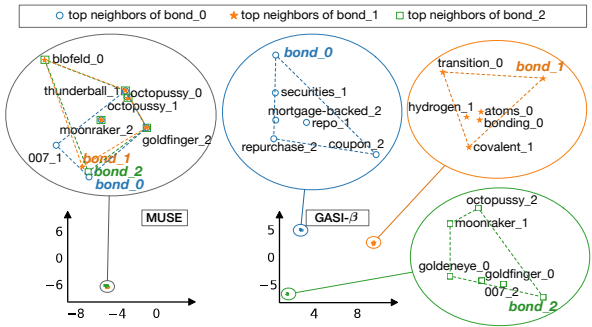}
	\ignore{-0.8cm}
	\caption{t-\abr{sne} projections of nearest neighbors for ``bond''
		by \emph{hard-attention} models: 
		\abr{muse} (\abr{rl}-based) and our 
		\gasi{-$\beta$}.
                Trained on same dataset and vocabulary,
		both models learn three vectors per word (bond$\_i$ is
		$i$\textsuperscript{th} sense vector). \gasi{} (right)
		learns three distinct senses of ``bond'' while \abr{muse}
		(left) learns overlapping senses.}
	\label{fig:tsne}
	\ignore{-0.5cm}
\end{figure}

Gumbel softmax learns a flat distribution over senses even with low
temperatures: the dot product $\bar{\bvec{c}}_i^\top \bvec{s}_k^i$ is
too small\footnote{This is from float32 precision and saturation of
  $\log(\sigma(\cdot))$\appendixmention{; detailed further in Figure~\ref{fig:dot} in
  Appendix}.} compared to the Gumbel noise $g_k$.  Thus we use a
scaling factor $\beta$ to encourage sparser
distributions,\footnote{Normalizing $\bar{\bvec{c}}_i^\top
  \bvec{s}_k^i$ or directly using $\log P(s_k^i \g w_i, \tilde{c}_i)$
  results in a similar outcome.}
\begin{equation}
\gamma_k^i = \softmax ((\bar{\boldsymbol{c}_i}^{\top}\boldsymbol{s}_{k}^i  + \beta g_k)/\tau),
\label{eq:scale}
\end{equation}
and tune it as a
hyperparameter.
We append \gasi{-$\beta$} to the name of models with a scaling
factor.
This is critical for learning
\emph{distinguishable senses} (Figure~\ref{fig:tsne},
Table~\ref{tab:wsim}, and Table~\ref{tab:human}). Our \textbf{final objective function} for \gasi{-$\beta$} is

\begin{equation}
J(\bmat{S}, \bmat{C}) \propto \sum_{w_i \in V}\sum_{w_{c}\in c_i}
\sum_{k=1}^{K} \gamma_k^i \log P(w_{c} \g s_k^i).
\label{eq:gasi}
\end{equation}

	\section{Data and Training}
\label{sec:setup}

\jbgcomment{This should not be its own section; integrate into model
  or move to supplement}

For fair comparisons, we try to remain consistent with previous
work~\citep{huang2012improving,neelakantan2015efficient,Muse} in all
aspects of training. In particular, we train \gasi{} on the same April
2010 Wikipedia snapshot~\citep{wiki} with 1B tokens and the same
vocabulary released by \newcite{neelakantan2015efficient}; set the
number of senses $K=3$ and dimension $d=300$ for each word unless
otherwise specified. \appendixmention{More details are in the
  Appendix.} Following \newcite{maddison2016concrete}, we fix the
temperature $\tau = 0.5$, and tune the scaling factor $\beta=0.4$
using grid search within $\{0.1\dots 0.9\}$ on \textbf{AvgSimC} for
contextual word similarity (Section~\ref{sec:wordsim}); this tuning
preceded all interpretability experiments. If not reprinted, numbers
for competing models are either computed with pre-trained embeddings
released by authors or trained on released code.

	\section{Evaluating Interpretability}
\label{sec:intp}

We turn to traditional evaluations of sense embeddings later
(Section~\ref{sec:wordsim}), but our focus is on human
interpretability.
If you
show a human the senses, can they understand why a model would assign
a sense to that context?
This section evaluates whether the
representations make sense to human consumers of multisense models.

In the age of \abr{bert} and \elmo{}, these are the dimensions that are most critical for
multisense representations.
While contextual word vectors are most useful for \emph{computer}
understanding of meaning, \emph{humans} often want an overview of word
meanings for other tasks.

Sense representations are useful for human-in-the-loop
applications.
They help understand semantic drift~\cite{hamilton-16}: how
do the meanings of ``gay'' reflect social progress?  They help people
learn languages~\cite{noraset-17}: what does it mean when someone says
that I ``embarrassed'' them?
They
help linguists understand the sense inventory of a
language~\cite{kawahara-14}: what are the frames that can be used by
the verb ``participate''?
These questions (and human understanding)
are helped by {\bf discrete senses}, which the Gumbel softmax uncovers.

More broadly, this is the goal of interpretable machine
learning~\cite{doshi-velez-17}.
While downstream models do not always
need an interpretable explanation of \emph{why} a model uses a particular
representation, interactive machine learning and explainable machine
learning do.
To date, multisense representations ignore this use case.

\begin{table}[t!]
	\centering
	\small
	\begin{tabular}{cccc}
		\toprule
		\multirow{2}{*} {Model} &  Sense  & Judgment & \multirow{2}{*} {Agreement}  \\
		&  Accuracy  & Accuracy&  \\
		\midrule
		\midrule
		\abr{muse}  & 67.33 & 62.89 & 0.73 \\
		\abr{mssg-30k}  & 69.33 & 66.67 & 0.76 \\
		\gasi{-$\beta$} & \textbf{71.33} &\textbf{ 67.33}& \textbf{0.77}\\
		\bottomrule
	\end{tabular} 
	\caption{Word intrusion evaluations on top ten nearest neighbors of sense embeddings.  Users find misfit words most easily with \gasi{-$\beta$}, suggesting these representations are more interpretable.}
	\label{tab:intrusion}
	\ignore{-0.5cm}
\end{table}

\ignore{-0.2cm}
\paragraph{Qualitative analysis}

Previous papers use nearest neighbors of a few examples to qualitatively
argue that their models have captured meaningful senses of words.
We also give an example in Figure~\ref{fig:tsne}, which provides an
intuitive view on how the learned senses are clustered by visualizing
the nearest neighbors of word ``bond'' using t-\abr{sne}
projection~\citep{maaten2008visualizing}.
Our model (right) disentangles the three sense of ``bond'' clearly.

However, examples can be cherry-picked. This
problem bedeviled topic modeling until 
rigorous human evaluation was introduced~\citep{chang-09b}.
We adapt both
aspects of their evaluations: \emph{word
  intrusion}~\citep{schnabel-15} to evaluate whether individual senses
are coherent and \emph{topic intrusion}---rather sense intrusion in
this setting---to evaluate whether humans agree with models' sense
assignments \emph{in context}.
Using crowdsourced evaluations from
Figure-Eight, we compare our models with two previous
state-of-the-art sense embeddings models, i.e.,
\abr{mssg}~\citep{neelakantan2015efficient} and
\abr{muse}~\citep{Muse}.\footnote{\abr{mssg} has two settings; we run human
  evaluation with \abr{mssg}-30K which has higher correlation
  with MaxSimC on \abr{scws}.}

\begin{table}[t]
	\small
	\begin{tabular}{cccc}
		\toprule
		Model &  Accuracy & $P$ & Agreement \\
		\midrule
		\midrule
		\abr{muse} & 28.0 & 0.33 &0.68\\
		\abr{mssg-30K} &  44.5 & 0.37 &0.73 \\
		\gasi{ (no $\beta$)} &  33.8 & 0.33& 0.68\\
		\gasi{-$\beta$} &  \textbf{50.0} & \textbf{0.48}& \textbf{0.75}\\
		\midrule
		\gasi{-$\beta$}-pruned & \textbf{75.2} &\textbf{0.67}& \textbf{0.96}\\
		\bottomrule
	\end{tabular} 
	\centering 
	\caption{Human-model consistency on \emph{contextual word sense
			selection}; $P$ is the average probability
		assigned by the model to the human choices. \gasi{-$\beta$} is most consistent with crowdworkers. Reducing sense duplications by retraining our model with pruning mask
		improves significantly human-model agreement.}   
	\label{tab:human}
	\ignore{-0.5cm}
\end{table}

\subsection{Word Intrusion for Sense Coherence}
\label{sec:word_intrusion}

\citet{schnabel-15} suggest a ``good'' word embedding should have
coherent neighbors and evaluate coherence by \emph{word
  intrusion}.
They present crowdworkers four words: three are close
in embedding space while one of which is an ``intruder''.
If the
embedding makes sense, contributors will easily spot the word that
``does not belong''.

Similarly, we examine the coherence of ten nearest neighbors of senses
in the \emph{contextual word sense selection} task
(Section~\ref{sec:sense_selection}) and replace one neighbor with an
``intruder''.
We generate three
intruders for each sense and collect three judgments per intruder.
To account for variation in users and intruders, we count an instance
as ``correct'' if two or more crowdworkers correctly spot the
intruder.

\ignore{
\begin{figure}[t!]
		\includegraphics[width=0.95\linewidth]{\figfile{word_intrusion}}
		\caption{Word intrusion task prompt }
		\label{fig:word_intrusion}
\end{figure}
}

Like \newcite{chang-09b}, we want the ``intruder'' to be about
as frequent as the target but not too similar.
For sense~$s_i^{m}$ of word type~$w_i$, we randomly select
a word from the neighbors of \emph{another} sense $s_i^{n}$ of $w_i$.

All models have comparable model accuracy.  \gasi{-$\beta$} learns
senses that have the highest coherence while \abr{muse} learns
mixtures of senses (Table~\ref{tab:intrusion}).

We use the aggregated confidence score provided by Figure-Eight to
estimate the level of \textbf{inter-rater agreement} between multiple
contributors~\cite{f8-confidence}.
The agreement is high for all models and \gasi{-$\beta$} has the
highest agreement, suggesting that the senses learned by
\gasi{-$\beta$} are easier to interpret.

\subsection{Contextual Word Sense Selection}
\label{sec:sense_selection}

The previous task measures whether individual senses are
coherent.
Now we evaluate models' disambiguation of senses \emph{in context}.

\paragraph{Task Description} 

Given a target word in context, we ask a crowdworker to select which
sense group best fits the sentence.
Each sense group is described by its top ten distinct nearest
neighbors, and the sense group order is shuffled.

\ignore{
 \begin{figure}[t]
   \includegraphics[width=7.5cm]{\figfile{ui_bond}}
   
 	\caption{An example (target: \textit{bond}) of the
 			 \textit{contextual word sense selection} task; each
 			option contains top ten nearest neighbors of a sense
 			embedding learned by the model; senses are from \gasi{-$\beta$} (1.~007;
 			2.~chemical; 3.~financial).}\label{fig:ui_bond}
 \end{figure}
}

\ignore{-0.2cm} 
\paragraph{Data Collection}  

We select fifty nouns with five sentences from SemCor
3.0~\citep{miller1994using}.
We first filter all word types with fewer than ten sentences and
select the fifty most polysemous nouns from WordNet~\citep{wordnet2}
among the remaining senses.
For each noun, we randomly select five sentences.

\ignore{-0.2cm} 
\paragraph{Metrics} 

For each model, we collect three judgments for each question.
We consider a model correct if at least two crowdworkers select the
same sense as the model.

\begin{table}
	\small
	\begin{tabular}{ccccc}
		\toprule
		&  &\abr{muse}& \abr{mssg}& \gasi{-$\beta$}\\
		\midrule
		
		\multirow{2}{*} {\parbox{1.3cm}{\centering word \\overlap}} &agree& 4.78 & 0.39&1.52\\
		&disagree&  5.43 & 0.98&6.36 \\
		\midrule
		\multirow{2}{*} {\parbox{1.3cm}{\centering Glove\\ cosine  }} &agree&  0.86& 0.33& 0.36\\
		&disagree &  0.88 & 0.57& 0.81\\
		\bottomrule
	\end{tabular} 
	\centering 
	
	\caption{Similarities of human and model choices when they
		agree and disagree for two metrics: simple word overlap (top) and
		Glove cosine similarity (bottom).  Humans agree with the model when
		the senses are distinct.}
	
	\label{tab:dist} 
	\ignore{-0.2cm} 
\end{table}

\begin{table}[t!]
	\small
	\centering
	\rowcolors{2}{gray!25}{white}
	\begin{tabular}{cp{6.5cm}}
		\toprule
		\multicolumn{2}{p{7.5cm}}{The real \underline{question} is - how are those four years used and what is their value as training?}   \\
		\midrule
		{\bf s1}& hypothetical, unanswered, topic, answered, discussion, yes/no, answer, facts\\
		s2& toss-up, answers, guess, why, answer, trivia, caller, wondering, answering\\
		s3& argument, contentious, unresolved, concerning, matter, regarding, debated, legality\\
		\bottomrule
	\end{tabular} 
	\centering 
	\caption{A case where \abr{mssg} has low overlap but confuses raters (agreement 0.33); model chooses s1.  }
	\label{tab:mssg_ea} 
	
	\ignore{-0.3cm} 
\end{table}

\ignore{-0.2cm} 
\paragraph{Sense Disambiguation and Interpretability} 

If humans consistently pick the same sense as the model, they must
first understand the choices, thus implying the nearest neighbor words
were coherent.
Moreover, they also agree that among those senses, that sense was the
right choice for this token.
\gasi{-$\beta$} selections are most consistent with humans';
it has the highest accuracy and assigns the largest probability
assigned to the human choices (Table~\ref{tab:human}).
Thus,
\gasi{-$\beta$} produces sense embeddings that are both more
interpretable and distinguishable. \gasi{} without a scaling factor,
however, has low consistency and flat sense distribution.

\ignore{-0.2cm} 
\paragraph{Model Confidence}

However, some contexts are more ambiguous than others.
For fine-grained senses, best practice is to use graded sense
assignments~\cite{erk-13}.
Thus, we also show the model's probability of the top human choice; distributions
close to $\frac{1}{K}$ (0.33) suggest the model learns a distribution
that cannot disambiguate senses.
We consider granularity of senses further in \secref{sec:related}.

\ignore{-0.2cm} 
\paragraph{Inter-rater Agreement} 

We use the confidence score computed by Figure-Eight to estimate the
raters' agreement for this task. \gasi{-$\beta$} has
the highest human-model agreement, while both \abr{Muse} and \gasi{}
without scaling have the lowest.

\ignore{-0.2cm} 
\paragraph{Error Analysis}

Next, we explore why crowdworkers disagree with the model even though
the senses are interpretable (Table~\ref{tab:intrusion}).
Is it that the model has learned \emph{duplicate} senses that both the
users and model cannot distinguish (the senses are all bad or
identical) or is it that crowdworkers agree with each other but
\emph{disagree} with the model (the model selects bad senses)?

Two trends suggest duplicate senses cause disagreement both for humans
with models and humans with each other.
For two measures of sense similarity---simple word overlap and
\glove{} similarity---similarity is lower when users and models agree
(Table~\ref{tab:dist}).
Humans also agree with each other more.  For \gasi{}-$\beta$, pairs with
perfect agreement have a word overlap of around 2.5, while the senses
with lowest agreement have overlap around 5.5.

To reduce duplicated senses, we
retrain the model with pruning (Section~\ref{sec:sasi}, Equation~\ref{eq:dup}).
We remove a little more than one sense per type on average.
To maintain the original setting, for word types that have fewer than three
senses left, we compute the nearest neighbors to dummy senses 
represented by random embeddings.
Our model trained with pruning mask 
(\gasi-$\beta$-pruned) reaches very high inter-rater agreement and higher
human-model agreement than models with 
a fixed number of senses (Table~\ref{tab:human}, bottom). 


\section{Word Similarity Evaluation}
\label{sec:wordsim}

\gasi{} and \gasi{-$\beta$} are interpretable, but how do they fare on
standard word similarity tasks?

\paragraph{Contextual Word Similarity}
Tailored for sense embedding evaluation, Stanford Contextual Word
Similarities~\citep[\abr{scws}]{huang2012improving} has 2003 word
pairs tied to context sentences.
These tasks assign a pair of word types (e.g., ``green'' and ``buck'') a
similarity/relatedness score.
Moreover, both words in the pair have an associated context.
These contexts disambiguate homonymous and polysemous
word types and thus captures sense-specific similarity.
%
Thus, we use this dataset to tune our hyperparameters, comparing
Spearman's rank correlation $\rho$ between
embedding similarity and the gold similarity judgments: higher
scores imply the model captures semantic similarities consistent with
the trusted similarity scores.

To compute the word similarity with senses we use two
metrics~\cite{Reisinger2010} that take context and sense
disambiguation into account:
\textbf{MaxSimC} computes the cosine similarity $\cos(s_1^*, s_2^*)$
between the two most probable senses $s_1^*$ and $s_2^*$ that
maximizes $P(s_k^i \g w_i, \tilde{c}_i)$.
\textbf{AvgSimC} weights average similarity over the combinations of
all senses $\sum_{i=1}^{K}\sum_{i=j}^{K} P(s_{i}^1 \g w_1,
\tilde{c}_1) P(s_{j}^2\g w_2,\tilde{c}_2)\cos(s_i^1 s_j^2)$.

\smallskip
We compare variants of our model with existing sense embedding
models (Table~\ref{tab:wsim}), including two previous \abr{sota}s: 
the clustering-based Multi-Sense Skip-Gram
model~\cite[\abr{mssg}]{neelakantan2015efficient} on AvgSimC 
and the \abr{rl}-based Modularizing Unsupervised Sense
Embeddings~\cite[\abr{muse}]{Muse} on MaxSimC.
\gasi{} better captures similarity than \sasi{},
corroborating that hard attention aids word sense selection.
\gasi{}
without scaling has the best MaxSimC; however, it learns a
flat sense distribution (Figure~\ref{fig:tsne}).
\gasi{-$\beta$} has
the best AvgSimC and a competitive MaxSimC.
While \abr{muse} has a
higher MaxSimC than \gasi{-$\beta$}, it fails to distinguish senses as
well (Figure~\ref{fig:tsne}, Section~\ref{sec:intp}).

We also evaluate the retrained model with pruning mask on this dataset.
\gasi-$\beta$-pruned has the same AvgSimC as \gasi-$\beta$ and higher
local similarity correlation (Table~\ref{tab:wsim}, bottom), validating our
pruning strategy (Section~\ref{sec:sasi}).

\begin{table}[t]
	\small
	\centering
	\begin{tabular}{ccc}
		\toprule
		Model  & MaxSimC & AvgSimC\\
		\midrule
		\midrule
		\newcite{huang2012improving}-50d & 26.1 & 65.7 \\
		\abr{mssg-6k}  &57.3 & 69.3 \\
		\abr{mssg-30k}  &59.3 & 69.2 \\
		\newcite{tian2014probabilistic} & 63.6 & 65.4 \\
		\newcite{li2015multi}& 66.6 &66.8 \\
		\newcite{qiu2016context} &64.9 & 66.1 \\
		\newcite{bartunov2016breaking} & 53.8 & 61.2\\
		\abr{muse}\_Boltzmann & 67.9 & 68.7 \\
		\midrule
		\sasi & 55.1 & 67.8 \\
		\gasi{} (w/o scaling) & \textbf{68.2}& 68.3\\
		\gasi-$\beta$  & 66.4& \textbf{69.5}\\
		\midrule
		\gasi-$\beta$-pruned  & 67.0& \textbf{69.5}\\
		\bottomrule
	\end{tabular}
    \ignore{-0.2cm} 
	\caption{Spearman's correlation $100\rho$ on \abr{scws} (trained on 1B token, 300d vectors except for Huang et al.).  \gasi{} and \gasi-$\beta$ both can disambiguate the sense and correlate with human ratings. Retraining the model with pruned senses furthur improves local similarity correlation.}
	
	\ignore{-0.5cm} 
	\label{tab:wsim} 
\end{table}

\ignore{-0.3cm} 
\paragraph{Word Sense Selection in Context} 

\abr{scws} evaluates models' sense selection indirectly. We
further compare \gasi{-$\beta$} with previous \abr{sota},
\abr{mssg-30k} and \abr{muse}, on the Word in Context
dataset~\citep[\wic{}]{pilehvar2018wic} which requires the model to
identify whether a word has the same sense in two contexts.
To reduce the variance in
training and to focus on evaluating the sense selection module, we use
an evaluation suited for unsupervised models: if the model selects
different sense vectors given contexts, we mark that the word has different
senses.\footnote{For monosemous or out of vocab words, we choose randomly.}
For \abr{muse}, \abr{mssg} and \gasi{-$\beta$}, 
we use each model's sense selection module; for DeConf~\citep{pilehvar2016} and \abr{sw2v}~\citep{sw2v}, we follow \citet{pilehvar2018wic} and \citet{makesense} by
selecting the closest sense vectors to the context vector.
DeConf results
are comparable to supervised results (59.4$\pm$ 0.7).
\gasi{-$\beta$} has the best result (55.3) apart from DeConf itself
(58.55)\appendixmention{(full results in Table~\ref{tab:wic} in
  appendix)}, which uses the same sense
inventory~\citep[WordNet]{wordnet2} as \wic{}.

\ignore{-0.3cm} 
\paragraph{Non-Contextual Word Similarity}

While contextual word similarity is best suited for our model and
goals, other datasets without contexts (i.e., only word
pairs and a rating) are both larger and ubiquitous for word vector
evaluations.
To evaluate the semantics captured by each sense-specific embeddings, we compare the models on non-contextual word similarity
datasets.\footnote{RG-65~\citep{rubenstein1965contextual};
SimLex-999~\citep{hill2015simlex};
WS-353~\citep{finkelstein2002placing};
MEN-3k~\citep{bruni2014multimodal};
MC-30~\citep{miller1991contextual}; YP-130~\citep{yang2006verb};
MTurk-287~\citep{radinsky2011word}; MTurk-771~\citep{halawi2012large};
RW-2k~\citep{luong2013better}}
Like \newcite{Muse} and \newcite{fastPG18}, we compute the word
similarity based on senses by \textbf{MaxSim}~\citep{Reisinger2010},
which maximizes the cosine similarity over the combination of all
sense pairs and does not require local contexts,
\begin{equation}
\text{MaxSim}(w_1, w_2) = \max_{0\le i \le K, 0\le j \le K}  \cos(s_i^1, s_j^2).
\label{eq:maxsim}
\end{equation}
 
\gasi{-$\beta$} has better correlation on three datasets, is
competitive on the rest (Table~\ref{tab:wsim2}), and remains
competitive without scaling.
\gasi{} is better than \abr{muse}, the
other hard-attention multi-prototype model, on six datasets and worse
on three. Our model can reproduce word similarities as well or better
than existing models through our sense selection.\footnote{Given how good \abr{pdf-gm} is, it could do better on contextual word similarity even though it ignores senses.  Average and MaxSim are equivalent for this model; it ties \gasi{-$\beta$}.}

\begin{table}[t]
  \small
	\centering
	\begin{tabular}{cccccccc}
		\toprule
		Dataset & \abr{muse} &\sasi &\gasi & \gasi{-$\beta$} & PFT-GM \\
		\midrule
		\midrule
		SimLex-999 & 39.61&31.56&40.14&\textbf{41.68}&40.19 \\
		WS-353& 68.41&58.31&68.49 &\textbf{69.36}&68.6 \\
		MEN-3k& 74.06 &65.07&73.13&72.32&\textbf{77.40} \\
		MC-30& 81.80&70.81&82.47 &\textbf{85.27}&74.63 \\
		RG-65& \textbf{81.11}&74.38&77.19&79.77&79.75 \\
		YP-130& 43.56&48.28&49.82&56.34&\textbf{59.39} \\
		MT-287& 67.22&64.54&67.37&66.13&\textbf{69.66} \\
		MT-771& 64.00&55.00&66.65&66.70&\textbf{68.91} \\
		RW-2k& \textbf{48.46}&45.03&47.22&47.69&45.69 \\
		\bottomrule
	\end{tabular}
    \ignore{-0.3cm} 
	\caption{Spearman's correlation on non-contextual word
          similarity (MaxSim). \gasi{-$\beta$} has higher correlation
          on three datasets and is competitive on the
          others. \appendixmention{\abr{pft-gm} is trained with two
            components/senses while other models learn three.  A full
            version including \abr{mssg} is in appendix.}}
	\label{tab:wsim2} 
	\ignore{-0.5cm}
\end{table}

\subsection{Word Similarity vs. Interpretability}

Word similarity tasks (Section~\ref{sec:wordsim}) and human
evaluations (Section~\ref{sec:intp}) are inconsistent.
\gasi{}, \gasi{-$\beta$} and \abr{muse} are
all competitive in word similarity (Table~\ref{tab:wsim} and
Table~\ref{tab:wsim2}), but only \gasi{-$\beta$} also does well in the
human evaluations (Table~\ref{tab:human}). Both \gasi{} without
scaling and \abr{muse} fail to learn distinguishable senses and cannot
disambiguate senses.
High word similarities do not necessarily indicate ``good'' sense
embeddings quality; our human evaluation---\emph{contextual word sense
  selection}---is complementary.


	\ignore{-0.1cm} 
\section{Related Work: Representation, Evaluation}
\label{sec:related}

\newcite{schutze1998automatic} introduces context-group discrimination
for senses and uses the centroid of context vectors as a sense
representation. Other work induces senses by context
clustering~\citep{purandare2004word} or probabilistic mixture
models~\citep{brody2009bayesian}. \newcite{Reisinger2010} first
introduce multiple sense-specific vectors for each word, inspiring
other multi-prototype sense embedding models. Generally, to address
polysemy in word embeddings, previous work trains on annotated
sense corpora~\citep{sensEmbed} or external sense
inventories~\citep{remb13,chen2014unified,jauhar2015ontologically,taoSense,wu2015sense,pilehvar2016,sw2v};
\newcite{autoE17}
extend word embeddings to lexical resources without training; others
induce senses via multilingual parallel
corpora~\citep{guo2014learning,vsuster2016bilingual,ettinger2016retrofitting}.

We contrast our \gasi{} to unsupervised monolingual multi-prototype models
along two dimensions: \emph{sense induction methodology} and
\emph{differentiability}.

\jbgcomment{This is just a big list.  Tell a story; where is the field going and what has it learned?}

On the dimension of \emph{sense induction methodology}, \newcite{huang2012improving} and \newcite{neelakantan2015efficient}
induce senses by context clustering; \newcite{tian2014probabilistic}
model a corpus-level sense distribution; \newcite{li2015multi} model
the sense assignment as a Chinese Restaurant Process;
\newcite{qiu2016context} induce senses by minimizing an energy
function on a context-depend network; \newcite{bartunov2016breaking}
model the sense assignment as a steak-breaking process; \newcite{mswe17}
model the sense embeddings as a weighted combination of topic vectors
with pre-computed weights by topic models; \newcite{fastPG18} model word representations as Gaussian Mixture
embeddings where each Gaussian component captures different
senses; \newcite{Muse} compute sense distribution by a separate set
of sense induction vectors. The proposed \gasi{} marginalizes the
likelihood of contexts over senses and induces senses by local context
vectors; the most similar sense selection module is a bilingual
model~\citep{vsuster2016bilingual} except that it does not introduce
lower bound for negative sampling but uses weighted embeddings, 
which results in mixed senses.

On the dimension of \emph{differentiability}, most sense selection
models are \emph{non-differentiable} and discretely select senses,
with two exceptions: \newcite{vsuster2016bilingual} use weighted
vectors over senses;
\newcite{Muse} implement hard attention with \abr{rl} to mitigate the non-differentiability.
In contrast, \gasi{} keeps full differentiability by
reparameterization and approximates discrete sense sampling with the
scaled Gumbel softmax.

However, the elephants in the room are \abr{bert} and \elmo{}.
While there are specific applications where humans might be better
served by multisense embeddings, computers seem to be consistently
better served by contextual representations.
A natural extension is to use the aggregate
representations of word senses from these models.
Particularly for
\elmo{}, one could cluster individual mentions~\cite{chang-19}, but
this is unsatisfying at first blush: it creates clusters 
more specific than senses.
\abr{bert} is even more
difficult: the transformer is a dense, rich representation, but
only a small subset describes the meaning of individual words.
Probing techniques~\cite{perone2018evaluation} could help focus on
semantic aspects that help \emph{humans} understand word usage.

\jbgcomment{Might be good to add some Bertology papers here that talk about all of the crazy stuff that gets stuffed into Bert vectors.}

\ignore{-0.1cm} 
\subsection{Granularity}

Despite the confluence of goals, there has been a disappointing lack
of cross-fertilization between the traditional knowledge-based lexical
semantics community and the representation-learning community.
We, following the trends of sense learning models, have---from the
perspective of those used to VerbNet or WordNet---used far too few
senses per word.
While there is disagreement about sense inventory, ``hard'' and
``line''~\cite{leacock-98} definitely have more than three senses.
Expanding to granular senses presents both challenges and
opportunities for future work.

While moving to a richer sense inventory is valuable future work, it
makes human annotation more difficult~\cite{erk-13}---while we can
expect humans to agree on which of three senses are used, we cannot for
larger sense inventories.
In topic models, \citet{chang-09b} develop topic log odds (in addition
to the more widely used model precision) to account for graded
assignment to topics.
Richer user models would need to capture these more difficult
decisions.

However, moving to more granular senses requires richer modeling.
Bayesian nonparametrics~\citep{orbanz-10} can determine the
number of clusters that best explain the data.
Combining online stick breaking distributions~\cite{wang-11} with
\gasi{}'s objective function could remove unneeded complexity for
word types with few senses and consider the richer sense inventory for
other words.

	\ignore{-0.3cm} 
\section{Conclusion}

The goal of multi-sense word embeddings is not just to win word sense
evaluation datasets. Rather, they should also \emph{describe}
language: given millions of tokens of a language, what are the
patterns in the language that can help a lexicographer or linguist in
day-to-day tasks like building dictionaries or understanding semantic
drift.  Our differentiable Gumbel Attention Sense Induction (\gasi{})
offers comparable word similarities with multisense representations
while also learning more distinguishable, interpretable senses.  

However, simply asking whether word senses look good is only a first
step.
A sense induction model designed for human use should be closely
integrated into that task.
While we use a Word2Vec-based objective function in
Section~\ref{sec:model}, ideally we should use a human-driven,
task-specific metric~\citep{feng-19} to guide the selection of senses
that are distinguishable, interpretable, {\bf and useful}.
\ignore{-0.2cm}



\bibliographystyle{style/acl_natbib}
\bibliography{bib/journal-full,bib/jbg,bib/fenfei,bib/miyyer,bib/fs}

\appendixmention{
\clearpage

\section*{Appendix}

Equation and
figure numbers continue from main submission.

\setcounter{section}{0}
\renewcommand{\thesection}{\Alph{section}}

\section{Derivation Desiderata}
\label{sec:deriv-dess}

Like the Skip-Gram objective (Equation~\ref{eq:sg_softmax}), we model
the likelihood of a context word given the center
sense~$P(c_j^i \g s_k^i)$ using softmax,
\begin{equation}
P(c_j^i \g s_k^i) = \frac{\exp \left({\bvec{c}_j^i} ^\top \bvec{s}_k^i\right)}{\sum_{j=1}^{|V|}\exp \left(\bvec{c}_j ^\top \bvec{s}_k^i\right)},
\end{equation}
where the bold symbol $\bvec{s}_k^i$ is the vector representation of sense
$s_k^j$ from $\bmat{S}$, and $\bvec{c}_j$ is the context
embedding of word $c_j$ from $\bmat{C}$.


Computing the softmax over the vocabulary is time-consuming. We want
to adopt negative sampling to approximate $\log P(c_{j}^i \g s_k^i)$,
which does not exist explicitly in our objective function
(Equation~\ref{eq:jatt}).\footnote{Deriving the negative sampling
  requires the logarithm of a softmax~\citep{goldberg2014negative}. }

However, given the concavity of the logarithm function, we can apply
Jensen's inequality,
\begin{align}
  \log & \left[ \sum_{k=1}^{K} P(c_{j}^i \g s_k^i) P(s_k^i \g w_{i}, \tilde{c}_i) \right] \ge \\
  & \sum_{k=1}^{K} P(s_k^i \g w_{i}, \tilde{c}_i) \log  P(c_{j}^i  \g
s_k^i),  \notag
\label{eq:jensen}
\end{align}
and create a lower bound of the objective.  Maximizing this lower
bound gives us a \emph{tractable objective}, $J(\bmat{S}, \bmat{C}) \propto$
\begin{equation}
\sum_{w_i \in V}\sum_{c_{j}^i \in \tilde{c}_i} \sum_{k=1}^{K}
P(s_k^i \g w_{i}, \tilde{c}_i) \log P(c_{j}^i  \g s_k^i),
\label{eq:att}
\end{equation}
where $\log P(c_{j}^i \g s_k^i)$ is estimated by negative
sampling~\cite{mikolov2013b},
\begin{equation*}
\text{log }\sigma({\bvec{c}_j^i}^{\top} \bvec{s}_k^i)  + \sum_{j=1}^{n}\mathbb{E}_{c_j\sim P_n(c)}[\text{log }\sigma(-\bvec{c}_j^{\top}\bvec{s}_k^j))]
\label{eq:neg}
\end{equation*}

\section{Training Details}
\label{apdx:train}

During training, we fix the window size to five and the dimensionality
of the embedding space to 300 for comparison to previous work. We
initialize both sense and context embeddings randomly within
U(-0.5/dim, 0.5/dim) as in Word2Vec. We set the initial learning rate
to 0.01; it is decreased linearly until training concludes after 5
epochs. The batch size is 512, and we use five negative samples per
center word-context pair as suggested by~\newcite{mikolov2013a}. The
subsample threshold is 1e-4. We train our model on the GeForce GTX
1080 Ti, and our implementation (using pytorch 3.0) takes $\sim6$
hours to train one epoch on the April 2010 Wikipedia
snapshot~\cite{wiki} with 100k vocabulary. For comparison, our
implementation of Skip-Gram on the same framework takes $\sim2$ hours
each epoch.

\begin{figure}[t]
	\centering
           \includegraphics[width=0.7\linewidth]{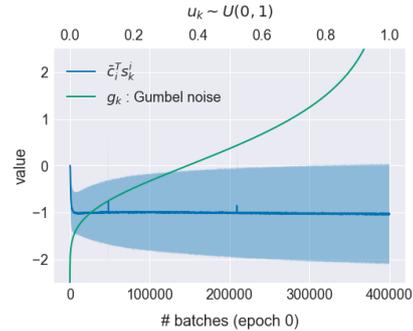}
           \ignore{-0.5cm}
           \caption{Our hard attention mechanism is approximated with
               Gumbel softmax on the context-sense dot product
               $\bar{\bvec{c}}_i^\top\bvec{s}_k^i$
               (Equation~\ref{eq:gs}), whose mean and std plotted here
               as a function of iteration. The shadowed area shows
               that it has a smaller scale than the Gumbel noise
               $g_k$, such that $g_k$, rather than the embeddings,
               dominates the sense attention.}\label{fig:dot}
\end{figure}

\begin{figure}[t]
	\centering
	\includegraphics[width=0.7\linewidth]{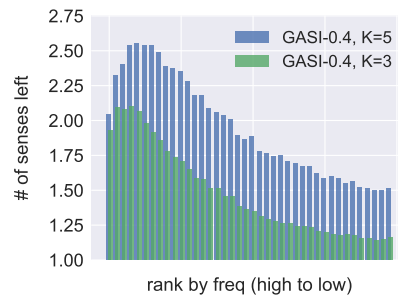}
	\ignore{-0.5cm}
	\caption{Histogram of number of senses left after post-training pruning for two models: \abr{gasi}-0.4 initialized with three senses and \abr{gasi}-0.4 initialized with five senses. We rank the number of senses of words by their frequency from high to low. }
	\label{fig:ns}
	\ignore{-0.5cm}
\end{figure}

\begin{figure*}[t]
	\centering
	\includegraphics[width=0.8\linewidth]{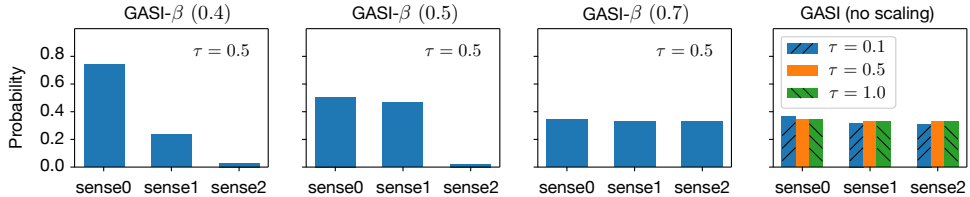}
	\ignore{-0.5cm}
	\caption{As the scale factor $\beta$ increases, the sense
		selection distribution for ``bond'' given examples from
		SemCor 3.0 for synset ``bond.n.02'' becomes flatter,
		indicating less disambiguated sense vectors.}
	\label{fig:bond}
\end{figure*}

\begin{table*}[t]
	\small
	\centering
	\begin{tabular}{cccccccc}
		\toprule
		Dataset &  \abr{mssg-30k}&  \abr{mssg-6k} &\abr{muse}\_Boltzmann &\sasi &\gasi & \gasi{-$\beta$} & PFT-GM \\
		\midrule
		\midrule
		SimLex-999 & 31.80 &28.65& 39.61&31.56&40.14&\textbf{41.68}&40.19 \\
		WS-353&65.69& 67.42 & 68.41&58.31&68.49 &\textbf{69.36}&68.6 \\
		MEN-3k& 65.99 & 67.10 & 74.06 &65.07&73.13&72.32&\textbf{77.40} \\
		MC-30& 67.79& 76.02&81.80&70.81&82.47 &\textbf{85.27}&74.63 \\
		RG-65& 73.90 & 64.97&\textbf{81.11}&74.38&77.19&79.77&79.75 \\
		YP-130& 40.69 & 42.68&43.56&48.28&49.82&56.34&\textbf{59.39} \\
		MT-287& 65.47& 64.04&67.22&64.54&67.37&66.13&\textbf{69.66} \\
		MT-771& 61.26& 58.83&64.00&55.00&66.65&66.70&\textbf{68.91} \\
		RW-2k& 42.87& 39.24&\textbf{48.46}&45.03&47.22&47.69&45.69 \\
		\bottomrule
	\end{tabular}
\ignore{-0.1cm}
	\caption{Spearman's correlation $100\rho$ on non-contextual
		word similarity (MaxSim). \gasi{-$\beta$}
		outperforms the other models on three datasets and is
		competitive on others. \abr{pft-gm} is trained with two components/senses while other models learn three.}
	\label{tab:wsim2-full} 
	\ignore{-0.3cm}
\end{table*}

\section{Number of Senses}
\label{apdx:sense}

For simplicity and consistency with most of previous work, we present
our model with a fixed number of senses~$K$.



\subsection{Post-training Pruning and Retraining}
\label{subsec:pruning}
For words that do not have multiple senses or have most senses appear
very low-frequently in corpus, our model (as well as many previous
models) learns duplicate senses.  Ideally, if we set a large number of $K$, with a perfect
\emph{pruning} strategy, we can estimate the number of senses per type by removing
duplicated senses and retrain a new model with the estimated number of senses instead 
of a fixed number $K$.

 However, this is challenging~\cite{mccarthy-16}; 
instead we use a simple pruning strategy and remove duplicated senses with a threshold
$\lambda$. Specifically, for each word $w_i$, if the cosine distance
between any of its sense embeddings ($\bvec{s}_{m}^i, \bvec{s}_{n}^i$)
is smaller than $\lambda$, we consider them to be duplicates. After
discovering all duplicate pairs, we start pruning with the sense
$s_k^i$ that has the most duplications and keep pruning with the same
strategy until no more duplicates remain.

\paragraph{Model-specific pruning} We estimate a model-specific threshold $\lambda$ from the learned embeddings instead of deciding it arbitrary. We first sample 100 words from the negative sampling distribution over the vocabulary. Then, we retrieve the top-5 nearest neighbors (from all senses of all words) to each sense of each sampled word. If one of a word's own senses appears as a nearest neighbor, we append the distance between them to a \emph{sense duplication list} $D_{dup}$. For other nearest neighbors, we append their distances to the \emph{word neighbor list} $D_{nn}$. After populating the two lists, we want to choose a threshold that would prune away all of the sense duplicates while differentiating sense duplications with other distinct neighbor words. Thus, we compute
\begin{equation}
\lambda = \frac{1}{2}(\mean(D_{dup}) + \mean(D_{nn})).
\end{equation}

\subsection{Number of Senses vs. Word Frequency}

It is a common assumption that more frequent words have more senses. Figure~\ref{fig:ns} shows a histogram of the number of senses left for words ranked by their frequency, and the results agree with the assumption. Generally, the model learns more sense for high frequent words, except for the most frequent ones. The most frequent words are usually considered stopwords, such as ``the'', ``a'' and ``our', which have only one common meaning. Moreover, we compare our model initialized with three senses (\gasi-0.4, $K=3$) against the one that has five (\gasi-0.4, $K=5$). Initializing with a larger number of senses, the model is able to uncover more senses for most words.

\subsection{Duplicated Senses and Human-Model Agreement}

We measure distinctness both by counting shared nearest neighbors and the average cosine similarities of
GloVe embeddings.\footnote{Different models learn different
  representations; we use GloVe for a uniform basis of comparison.}
Specifically, \abr{muse} learns duplicate senses for most words,
preventing users from choosing appropriate senses and preventing human-model agreement. \gasi{-$\beta$} learns some duplicated
senses and some distinguishable senses. \abr{mssg} appears to learn
the fewest duplicate senses, but they are not distinguishable enough for
humans. Users disagree with each
other (0.33 agreement) even when the number of overlaps is very small
(Figure~\ref{fig:dist_user}). Table~\ref{tab:mssg_ea} shows an intuitive example. If we use rater agreement to measure how distinguishable
the learned senses are to humans, \gasi{-$\beta$} learns the most
distinguishable senses.

The model is more likely to agree with humans when humans agree with
each other (Figure~\ref{fig:acc_user}), i.e., human-model consistency
correlates with rater agreement
(Figure~\ref{fig:acc_user}). \abr{mssg} disagrees with humans more
even when raters agree with each other, indicating worse sense
selection ability.

\begin{figure}[t]
		\includegraphics[width=0.8\linewidth]{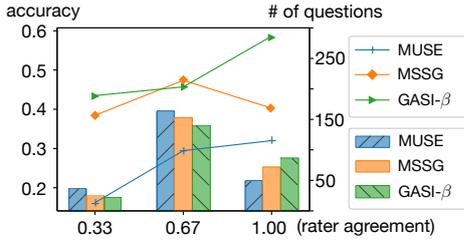}
		\ignore{-0.5cm}
		\caption{Higher inter-rater agreement correlates with higher human-model consistency. }
		\label{fig:acc_user}
\end{figure}

\begin{table}[t]
	\small
	\centering
	\begin{tabular}{cc}
		\toprule
		Model  & Accuracy(\%)\\
		\midrule
		\multicolumn{2}{l}{\textit{unsupervised multi-prototype models}}\smallskip\\
		\abr{mssg-30k}  &54.00  \\
		\abr{muse}\_Boltzmann & 52.14 \\
		\gasi-$\beta$  & \textbf{55.27}\\
		\midrule
		\multicolumn{2}{l}{\textit{semi-supervised with lexical resources}}\smallskip\\
		DeConf & \textbf{58.55}\\
		\abr{sw2v} & 54.56 \\
		\bottomrule
	\end{tabular}
\ignore{-0.2cm}
	\caption{Sense selection on Word in Context (\wic{}) dataset.}
	\label{tab:wic} 
	\ignore{-0.2cm}
\end{table}

\begin{figure}[t]
	\centering
	\includegraphics[width=0.7\linewidth]{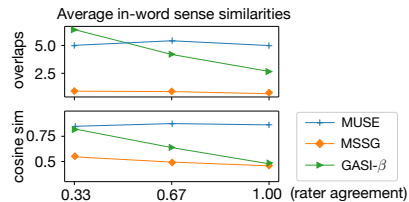}
	\ignore{-0.4cm} 
	\caption{Overall, human agree more with each other when the senses are more distinct (less word overlaps and smaller cosine similarities)}
	\ignore{-0.5cm} 
	\label{fig:dist_user}
	
\end{figure}

}

\end{document}